\documentclass[a4paper, 10pt, conference]{ieeeconf}      

\IEEEoverridecommandlockouts                              

\usepackage[utf8]{inputenc}
\usepackage[whole]{bxcjkjatype} 
\usepackage[pdftex]{graphicx}
\usepackage[hyphens]{url}

\title{\LARGE \bf
 ditlab system for Dialogue Robot Competition 2022
}

\author{Yuuki Tachioka$^{1}$ 
\thanks{*This system had been constructed for the Dialogue Robot Competition 2021 with Atsushi Keyaki who worked at Denso IT Laboraory}
\thanks{$^{1}$Denso IT Laboratory,
        2-15-1 Shibuya, Shibuya, Tokyo, Japan
        {\tt\small tachioka.yuki@core.d-itlab.co.jp}}%
}

\begin{document}

\maketitle
\thispagestyle{empty}
\pagestyle{empty}

\begin{abstract}
We developed a dialogue system for Dialogue Robot Competition 2022.
Our system is composed of three parts.
First part investigates participants' demographic information by rule-based interview.
Second part recommends a point of interest (POI) based on the collected demographic information.
Third part answers participants' question based on the combination of rule-based answering and deep-learning-based answering with nearby POI search.
\end{abstract}

\section{INTRODUCTION}

Dialogue Robot Competition 2022 is a travel agency dialogue task, which aims to develop a dialogue service with hospitality.
Detailed conditions are described in the papers published by the organizers \cite{Higashinaka2022,Minato2022}.
Our system honors customers' preference and assists customers' decision.
Types of point of interest (POI) are classified into sightseeing type or experience type.
Interview with customers obtains demographic information and determines which POI is preferable for the customer.
System recommends a preferable POI and explains a ground based on the demographic attributions or travel conditions.
System answers customers' question in two types of systems that pick up a corresponding pair of question and answer based on a keyword search or generate answers by a neural-based system.
In addition, to make the given information more attractive, nearby POIs are searched and POI with better reputation is recommended.

\section{DIALOGUE FLOW}

\subsection{Overview of system}

\begin{figure}[tb]
\begin{center}
\includegraphics[width=0.96\linewidth]{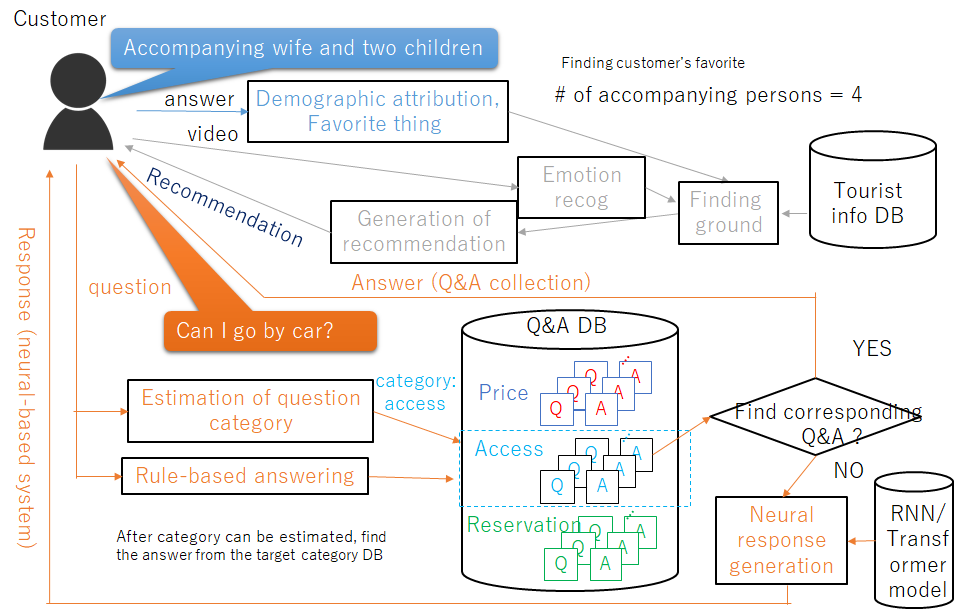}
\end{center}
\caption{Overview of our system}
\label{fig:overview}
\end{figure}

Fig.~\ref{fig:overview} shows an overview of the system composed of two elements.
First element makes a recommendation of POI considering customers' attribution of demographic information or preference estimated from customers' interviews.
To estimate customer's preference, system uses collected information or emotion recognition results.
Second element is a question and answer part.
Two types of methods are used to make answers: rule-based one or neural dialogue generation.
Instead of finding a corresponding question and answer from entire question and answer database, system collates a question corresponding to the target category after category estimation of a customers' question.
If appropriate answer cannot be found in question and answer pairs, neural dialogue generation can generate an answer.
In addition, nearby POI is searched, if a customer is interested.

\subsection{Investigation of demographic information}

Demographic information is extracted from an interview with a customer.
First, our system asks a name of customers to make the customer feel familiar with the system.
To avoid misunderstanding, if answer is recognized as famous family names (top 5,000 in Japan), the recognition result is adopted and this is used for calling customers.
Second, to clarify the customer's demographic information and preference, following questions are asked to the customers.
\begin{enumerate}
\item How many times did you visit Odaiba?
\item How many people are you accompanying with?
\item Which types of travel is favorite? (sightseeing type: watching exhibition as you like; experience type: experience something by yourself)
\item (In the case of experience type and if the number of accompanying person is 3 and more,) do you accompany small children?
\item (If recommended POIs can allow visitors to accompany pet,) do you intend to accompany pets?
\end{enumerate}

\begin{figure}[tb]
\begin{center}
\includegraphics[width=0.96\linewidth]{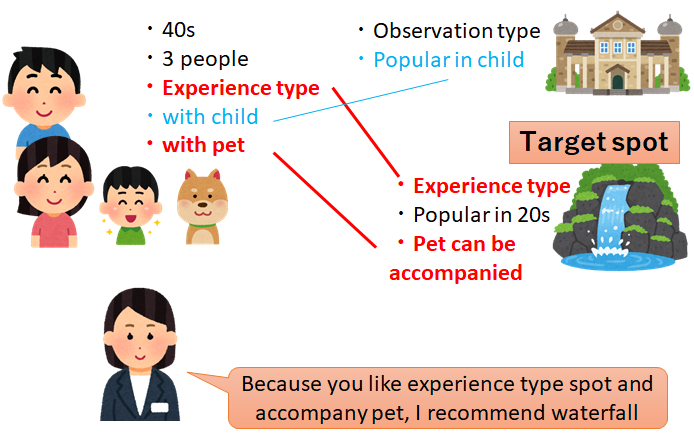}
\end{center}
\caption{Example of recommendation by customer's demographic information and  preference}
\label{fig:recommendation}
\end{figure}

\subsection{Recommendation}

Recommendation is made by grounds.
Fig.~\ref{fig:recommendation} shows an example of recommendation.
System finds grounds of recommendation from demographic attributions such as age, accompanying persons, and preferences.
There are two types of POIs: sightseeing type or experience type.
User's preferences are matched with a type of POI and the matched POI is recommended with grounds.
Information of POI is collected by web search (Jalan\footnote{\url{https://www.jalan.net/kankou/}} or Google Map)

When our system explains the information of POI and recommendation grounds, default speech synthesis is monotone and not natural.
Thus, our system emphasizes important words in the explanations.
Important words are specified by a calculation of BM25 (weighting of important words) \cite{Robertson2009}.
Top three important words are emphasized by slowing down the utterance speed.

\subsection{Question and answer (Estimation of question category and rule-based answering)}

\label{ss:rule_QA}

For a dialogue system that uses natural language, users ask various questions.
Thus, the direct collation of related question is difficult because to prepare those, the number of anticipated question and answer pairs is large.
It is effective to estimate a category of customers' questions before searching a pair of question and answer, because the estimation accuracy can be high if the category of the question is narrowed down.
Table~\ref{tab:category} shows fourteen categories to be classified.
Each question is classified into one category.
For classification, BERT \cite{Devlin2019} and Wikipedia2Vec \cite{Yamada2020} are employed.
For training, to construct pairs of question and  answers, the information of POI provided by organizers is extended by taking additional questions from frequently asked questions of websites.
For example, Are there restaurants nearby? (cafe, restaurant, and service) When is the closed day? (open hours) If I come by car, is there any parking? (access information) Is it possible to watch the exhibition in English? (information of exhibition and experience)
For some categories, there are small amount of data.
By using crowdsourcing, questions are collected.
System picks up appropriate questions from question collection that is related to the estimated question category.
After the pickup, corresponding answer can be found by using rule-based keyword search.

\begin{table}[tb]
\centering
\tabcolsep 1pt
\caption{Question categories}
\label{tab:category}
\begin{tabular}{c||c}
\hline
cafe, restaurant, and service & accessibility \\
museum shop & rules \\
assistance of education & information of institution \\
open hours & access information \\
nearby POI & equipment \\
group admission & information of exhibition and experience \\
reservation & price \\
\hline
\end{tabular}
\end{table}

\begin{table*}[tb]
\centering
\caption{Example of responses to ``先週は神戸に行ったよ (Last week I went to Kobe)''}
\label{tab:response}
\begin{tabular}{c|c}
\hline
Model & Response\\
\hline
RNN (Chiebukuro/slack/dialogue book) & 神戸には行ったことがありますか (Did you go to Kobe?) \\
Transformer1 (Chiebukuro/slack/dialogue book) & そうなんですね。私も神戸に行きました (Really? I also went to Kobe) \\
Transformer2 (NTT released data) & どうだった?楽しかった? (How? Did you enjoy it?) \\
RNN+Transformer1 & そうなんですね。私も神戸に行きました (I see. I went to Kobe) \\
RNN+Transformer\{1+2\} & そうなんですね。私も行ってみたいです (I see. I'd like to go to Kobe)\\
\hline
\end{tabular}
\end{table*}

\subsection{Question and answer (Deep-learning-based system)}

It is impossible to answer all questions in the way of \ref{ss:rule_QA}.
To answer questions for which we cannot prepare pairs of question and answer, deep-learning-based methods such as sequence-to-sequence (seq2seq) method \cite{Sutskever2014} have been used.
seq2seq model is a translation model from a user utterance to the system response \cite{Vinyals2015}.
OpenNMT \cite{opennmt} was used for answer generation.
It is possible to generate responses to any user utterances but these models require a lot of training data.
Transformer \cite{Vaswani2017} was used in addition to RNN encoder-decoder model.
First, we pre-trained models on open2ch and twitter datasets.
Second, we fine-tuned models on Yahoo! Chiebukuro, simulated dialogue by slack (internal collection) and dialogue data attached to a book \cite{Higashinaka2020}.
Third, we further fine-tuned transformer model on the dialogue data released from NTT \footnote{\url{https://github.com/nttcslab/japanese-dialog-transformers}} and question and answer pairs from ``AI王''(AI king)\footnote{\url{https://sites.google.com/view/project-aio/dataset}}.
RNN and transformer with two types of training data were combined to generate answers.

Table~\ref{tab:response} shows an example of response to ``Last week, I went to Kobe''.
RNN's responses are frequently similar to the customers' utterances.
Transformer (with data from NTT) generates various responses but too frank.
Combination can generate more reasonable responses.

\subsection{Search for recommended nearby point-of-interest}

Question about nearby POIs is processed by Google places API\footnote{\url{https://developers.google.com/maps/documentation/places/web-service/overview}}, which can collect the information of nearby POI within less than 800m.
If a customer specifies a genre of POI such as restaurant, cafe, or park, our system searches POI with the specified genre.
By using distance matrix API\footnote{\url{https://developers.google.com/maps/documentation/distance-matrix/overview}}, the distances on foot from the target POI to the found nearby POI are sorted in ascending order.
From the nearest one, our system collects reputations and introduces the highest score one to eliminate negative comments.
If the length of comments is too long, the first two sentences are extracted.
If customer did not ask any questions and dialogue time remained, even when customer did not ask questions about nearby POI, system introduces a nearby POI in some genres that have not been introduced.

\subsection{Robot control}

Facial expression and head movement of the android robot used in the competition can be controlled.
To give a customer a feeling of relief, the basic facial expression is smile and sometimes to give a feeling of tension, normal facial expression is used.
When a customer answers a question, the android nods.
When the robot explains the description of POI, she gazes at the photo of the monitor.
Because a customer's age is given by the organizer, the utterance speed is adjusted, faster for younger and slower for elder.

\section{CONCLUSION}

We developed a dialogue system that recommends a POI that matches to the user demographic attribution or preferences, which the system obtains via interviews with a customer.
To answer various questions from customers, answers are generated by rule-based keyword search and answer generation with a search for nearby POIs.
Facial expression, head movement, and utterance speed of the android robot are adjusted for natural and comfortable dialogue.


\end{document}